\PassOptionsToPackage{table}{xcolor}
\documentclass[sigconf]{acmart}
\usepackage{hyperref}

\usepackage{algorithm}
\usepackage{algorithmic}

\usepackage{adjustbox}
\usepackage{soul}

\usepackage{booktabs} 

\usepackage{pifont}
\usepackage{fancynum}
\usepackage{soul}
\usepackage{multirow}
\usepackage{subfigure}
\usepackage{float}

\usepackage{subfigure}
\usepackage{pifont}
\usepackage{tcolorbox}
\usepackage{balance}

\newcommand{\cmark}{\ding{51}}%
\newcommand{\xmark}{\ding{55}}%

\newcommand{\etal}{\textit{et al}.}
\newcommand{\ie}{\textit{i}.\textit{e}.}
\newcommand{\eg}{\textit{e}.\textit{g}.}
\newcommand{\vs}{\textit{v}\textit{s}.}

\newcommand{\modelname}{{DeVi}}
\newcommand{\datasetname}{{DeVE-QA}}

\copyrightyear{2025}
\acmYear{2025}
\setcopyright{acmlicensed}
\acmConference[SIGIR '25]{Proceedings of the 48th International ACM SIGIR
Conference on Research and Development in Information Retrieval}{July 13--18,
2025}{Padua, Italy}
\acmBooktitle{Proceedings of the 48th International ACM SIGIR Conference on
Research and Development in Information Retrieval (SIGIR '25), July 13--18,
2025, Padua, Italy}
\acmDOI{10.1145/3726302.3729945}
\acmISBN{979-8-4007-1592-1/2025/07}
\begin{document}

\title{Question-Answering Dense Video Events} 
\author{Hangyu Qin}
\affiliation{
  \institution{National University of Singapore}
  \city{Singapore}
  \country{Singapore}
}
\email{hqin@comp.nus.edu.sg}

\author{Junbin Xiao}
\authornote{Corresponding Author}
\affiliation{
  \institution{National University of Singapore}
  \city{Singapore}
  \country{Singapore}
}
\email{junbin@comp.nus.edu.sg}

\author{Angela Yao}
\affiliation{
  \institution{National University of Singapore}
  \city{Singapore}
  \country{Singapore}
}
\email{ayao@comp.nus.edu.sg}

\renewcommand{\shortauthors}{Hangyu Qin, Junbin Xiao, \& Angela Yao}

\begin{abstract}
This paper presents question-answering on dense video events, a novel task that answers and grounds dense-event questions in long videos, thus challenging MLLMs to faithfully comprehend and reason about multiple events over extended periods of time. To facilitate the study, we construct DeVE-QA -- a dataset featuring 78$K$ questions about 26$K$ events on 10.6$K$ long videos. Our benchmarking shows that state-of-the-art MLLMs struggle on DeVE-QA. For improvement, we propose DeVi, a novel training-free MLLM approach that highlights a hierarchical captioning module, a temporal event memory module, and a self-consistency checking module to respectively detect, contextualize and memorize, and ground dense-events in long videos for question answering. Extensive experiments show that DeVi is superior at answering dense-event questions and grounding relevant video moments. Compared with existing MLLMs, it achieves a notable increase of 4.8\% and 2.1\% for G(round)QA accuracy on DeVE-QA and NExT-GQA, respectively. Data and code are available at \textit{\href{https://github.com/QHUni/DeVE-QA}{https://github.com/QHUni/DeVE-QA}}. 
\end{abstract}




\maketitle

\section{Introduction}\label{sec:intro}
\begin{figure}[t]
\centering
\includegraphics[width=0.48\textwidth]{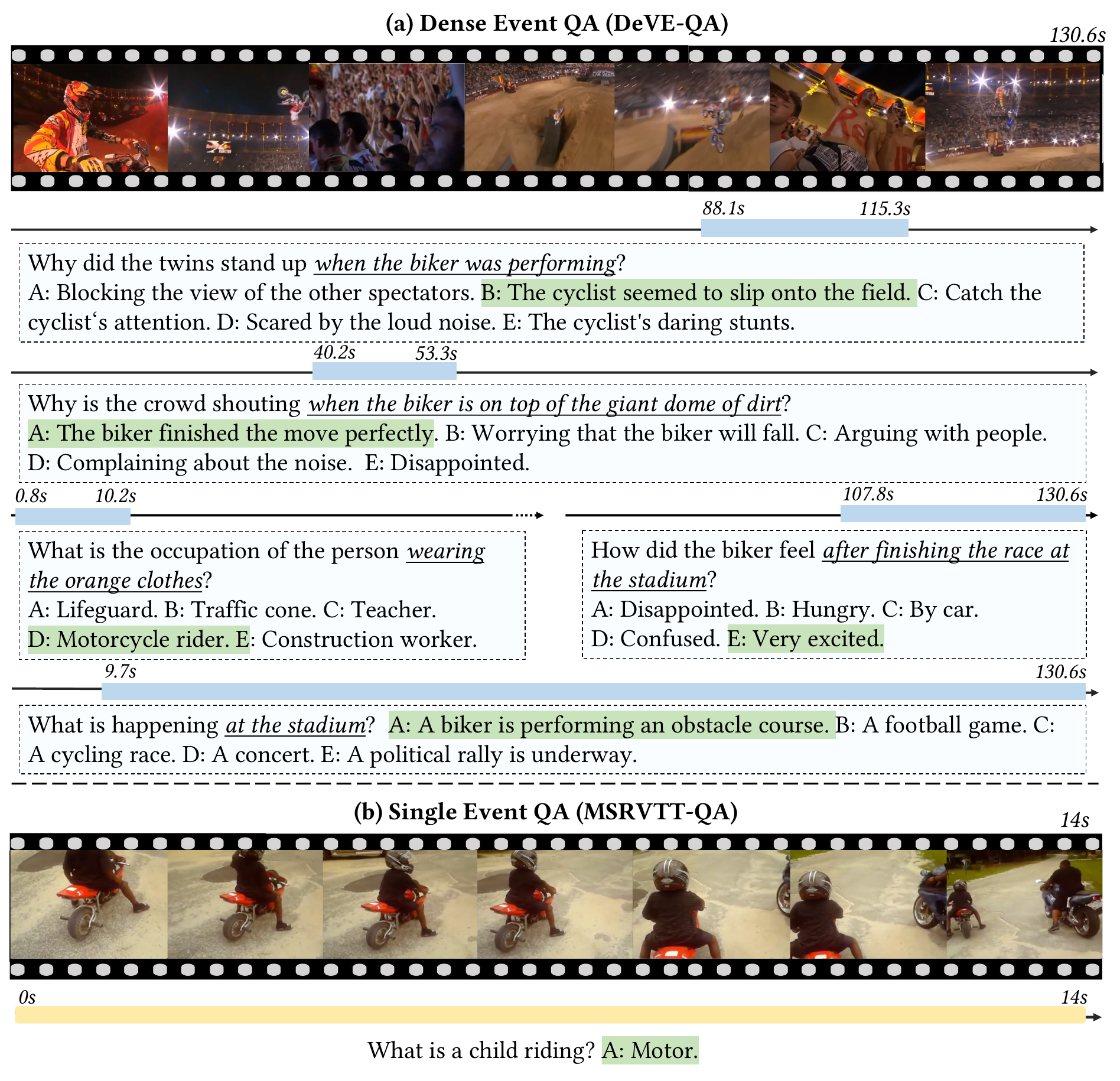} 
\vspace{-0.8cm}
\caption{Question answering dense video events on DeVE-QA (a) \vs~ single video event on MSRVTT-QA (b).}
\label{fig:intro}
\vspace{-0.4cm}
\end{figure}

Multimodal Large Language Models (MLLMs) \cite{alayrac2022flamingo,li2023videochat,maaz2023video,zhang2023video,lin2023video,gemini2} are highly capable at question-answering (QA) for single-event videos \cite{xu2017video,ye2017video,jang2017tgif}.  Such videos are short (3 $\sim$ 20 seconds) and the QAs factor single, global events, \eg~``\emph{who did what}''.
Yet, real-world videos are long and feature a complex overlay of \emph{dense} events.  
Consider the 2-minute video of a motorcycle activity shown in Figure~\ref{fig:intro} (Top).
A variety of questions can be asked about this video, each pertaining to an individual event but involving different participants and durations interspersed throughout the video. 

The inherent challenge of understanding dense video events is to isolate and or agglomerate, as needed, relevant video content and generate responses. Part of this challenge is tackled by dense captioning~\cite{krishna2017dense}, \ie, isolation and generation.  However, captioning is holistic and offers limited insight for reasoning of dense video events, as MLLMs are prone to hallucination~\cite{ma2023vista}. 
Furthermore, evaluating captions is challenging, as the annotations are often subjective~\cite{wang2024cycle} and the captions come in diverse language formats \cite{vedantam2015cider}. 
Alternatively, video question answering inherits all the challenges of dense event understanding. It also enables deterministic evaluation
by multi-choice classification~\cite{xiao2021next,mangalam2024egoschema,patraucean2024perception}.
As such, we propose question-answering of dense video events, a novel task that challenges MLLMs in   
comprehension and reasoning over dense events in long-form videos.

Given a video with multiple events and a question about one specific event, dense video QA requires MLLMs to comprehend and link the question to the relevant event, and reason over the event for the correct answer. To indicate comprehension, we require models to localize the relevant video moments \cite{xiao2023can,yang2021deconfounded} and substantiate the predictions with visual evidence. 
This requirement has three challenges.
\textbf{First}, each question pertains to a specific event, 
where events may vary in duration (see Figure~\ref{fig:intro}). 
It is therefore imperative to capture the events spanning over different time scales. 
\textbf{Second}, the long-form videos pose a challenge in relating possibly distant context for understanding a specific event. \textbf{Finally}, to promote faithful reasoning, a correct answer prediction necessitates correct grounding and question-answering. This asks for strong capability of dense visual event understanding and conditioning, versus exploiting common-sense knowledge in LLMs. 

Currently, there is no suitable benchmark for question-answering dense video events.  We construct \datasetname, a \textbf{\underline{De}}nse \textbf{\underline{V}}ideo \textbf{\underline{E}}vent \textbf{\underline{QA}} dataset featuring
$78K$ questions about 26$K$ events on $10.6K$ videos. 
\datasetname~is constructed by curating multi-choice questions from 
the dense-event caption annotations of ActivityNet-Caption \cite{krishna2017dense}.  Specifically, we prompt GPT-4 and then apply 
rigorous manual checking and corrections.  
Prominent MLLMs~\cite{wang2023internvid, yu2024self, momeni2023verbs, suris2023vipergpt, zhang2023simple, kim2024image, GPT-4o, zhang2024long, wang2024qwen2} that perform well in standard video QA struggle on \datasetname, especially on the subsets with denser events and longer videos, suggesting the significant challenge carried by \datasetname.

For improvement, we propose \modelname, a training-free MLLM approach for dense video-event QA. To solve the aforementioned challenges, we incorporate three specific strategies:
1) hierarchical dense event captioning to detect the dense events at multiple temporal scales, 2) temporal event contextualizing and memorizing to capture long-term event dependency and to facilitate event-grounded QA,  
and 3) self-consistency checking to anchor or rectify the answers with regard to the grounded event moments.
\modelname achieves accuracy increases of 4.8\% and 6.9\% over state-of-the-art 
on \datasetname~for QA with and without grounding, respectively. It also improves GQA accuracy on NExT-GQA~\cite{xiao2023can} by 2.1\%. Further ablation and in-depth analyses of different event densities and video lengths validate \modelname's strength and its particular designs for dense-event and long-form video QA. Additionally, we share our investigation of alternative implementations for \modelname, \eg~different MLLMs for captioners and QA models, and highlight the crucial importance of large models for success.

To summarize, our contributions are as follows:
\begin{itemize}
    \item We propose question-answering dense video events to challenge MLLMs in comprehending and reasoning the dense events in long videos and construct the \datasetname~dataset to facilitate the study.

   \item We propose \modelname, a training-free MLLM approach that performs grounded question-answering on dense video events with three dedicated components: hierarchical dense-event captioning, event contextualizing and memorizing, and self-consistency checking.
   
   \item We achieve new SoTA zero-shot results on both \datasetname~and NExT-GQA, surpassing the previous SoTAs profoundly by 4.8\% and 2.1\%, respectively. 
\end{itemize}
\vspace{-0.5cm}

\section{Related Works}

\subsection{Dense Event Video Understanding}
Dense video event understanding has primarily focused on captioning~\cite{krishna2017dense,wang2018video,lin2022swinbert,yang2023vid2seq}. 
However, optimizing for holistic sentence generation often results in several challenges. One common issue is overfitting~\cite{chen2020delving}, where models optimized for generating holistic sentences tend to perform well on training data but struggle with unseen data due to the complex variability in real-world events. Another issue is object hallucination~\cite{rohrbach2018object}, where models erroneously describe objects that are not present in the video, a phenomenon documented by Rohrbach~\etal~\cite{rohrbach2018object}. These challenges are exacerbated by the reliance on subjective caption annotations and the limitations of standard evaluation metrics like BLEU~\cite{papineni2002bleu} and CIDEr~\cite{vedantam2015cider}, which focus on sentence matching rather than video understanding.
In light of this, we opt for question-answering as an alternative task to evaluate the understanding and reasoning of dense video events.

\subsection{Video Question Answering}
Question answering serves as an important way to develop and test the visual understanding and reasoning capabilities of Vision-Language Models (VLMs). 
Popular VideoQA benchmarks (such as TGIF-QA~\cite{jang2017tgif}, Youtube2Text-QA \cite{ye2017video}, MSRVTT-QA and MSVD-QA~\cite{xu2017video}) and techniques (including recent MLLMs:  InstructBLIP~\cite{dai2023instructblip}, Video-ChatGPT~\cite{maaz2023video}, Video-LLaMA \cite{zhang2023video}, and VideoChat2~\cite{li2024mvbench}) focus on short videos and the questions primarily concern visual facts such as object attributes, locations, and actions \cite{zhong2022video}.
While ActivityNet-QA~\cite{yu2019activitynet} features long videos, its questions share the same content on visual element recognition.
These benchmarks and related techniques lack a specific focus on human-centered events and multi-event reasoning, especially in long-range videos. 

More recent VideoQA benchmarks~\cite{xiao2021next, choi2021dramaqa, fu2024video, zhou2024mlvu, suris2023vipergpt} emphasize either inferential queries about action relations or pure long video understanding.
NExT-QA~\cite{xiao2021next}, for example, advances beyond prior work by focusing on multiple action relations, but the videos remain centered on daily life scenarios without delving into the intricate understanding of multiple events in long-lasting activities. Video-MME \cite{fu2024video} and MLVU \cite{zhou2024mlvu} especially underscore long video understanding for MLLMs, neglecting dense events present in long videos. The closest benchmark to ours in task format is NExT-GQA \cite{xiao2023can}. Yet, its videos and QAs are sourced from NExT-QA, and thus do not challenge dense event understanding. For techniques, while some recent publications claim to address event-centric VideoQA \cite{yin2023cross, liu2023cross, bai2024glance}, the event in these works typically refers to an individual action or a global event, lacking the granularity required for understanding multiple and dense events in long videos. 

\subsection{MLLMs for VideoQA} 
Most existing Video-LLMs are designed for understanding short and single-event videos \cite{maaz2023video,zhang2023video,lin2023video,li2023videochat,li2024mvbench, yu2024self, ko2023large,kim2024image,xiao2024videoqa}. They typically accept inputs of 4 to 32 frames. 
This constraint prevents them from addressing complex reasoning over long-form videos, which are key for dense event understanding.
In an attempt to bypass the temporal constraints, recent research goes in two directions. The first is long video encoding, which explores token compression techniques to enable long video input. Typical works are LLaMA-VID \cite{li2024llamavid}, PLLaVA \cite{xu2024pllava}, LLaVA-NeXT-Video \cite{zhang2024llavanextvideo}, and LongVA \cite{zhang2024long}. The second is 
Scoratic technique \cite{zengsocratic}, which carefully composes pretrained foundation models in a zero-shot manner to exchange information with each other and capture new multimodal capabilities. Example solutions can be further classified into traversal (such as ViperGPT \cite{suris2023vipergpt}, MoReVQA \cite{min2024morevqa}, TraveLER \cite{shang2024traveler}, and VideoAgent \cite{fan2025videoagent,wang2025videoagent}) and vision-to-text memorizing (like LLoVi \cite{zhang2023simple}) ones. The traversal approaches cannot agglomerate multiple events interspersed at different times for joint reasoning.
Thus, we follow LLoVi~\cite{zhang2023simple} to convert frame sequences into time-aware descriptions and memorize them for question answering. 
To handle dense-event QA, we go beyond by incorporating specialized modules such as hierarchical captioning at multiple temporal scales, event contextualizing for long-range information interaction, and visual memorizing for self-verification.

\section{\datasetname~ Dataset}\label{sec:TD}

\subsection{Dataset Construction} \label{sec:DC}
We follow dense-event captioning \cite{krishna2017dense} to define an event as a completed description of a person's (or a group's) specific behavior within a specific time, \eg, ``\emph{A man is playing the piano at [10.2s, 34.5s]}''. Thus, we source our dataset \datasetname~ from dense event captioning dataset ActivityNet-Captions.  From these captions, we derive the corresponding question-answer pairs by prompting GPT-4 \cite{GPT-4} followed by human checking and correction. 
We elaborate on the construction details below.

\textbf{Data Filtering}. 
On average, the dense captions from ActivityNet-Captions~\cite{krishna2017dense} have a length of 13.5 words and span 94.6\% of the video. We first select samples that exceed the average caption length to focus on highly dense cases with more fine-grained visual event descriptions. Also, we omit the event samples that span more than 95\% of the corresponding video because the temporal moments of these events can be approached by simply returning the start and end time of the videos. 
Finally for efficiency, we perform a random sample over the remaining data to get a subset of 10,643 videos and 26,111 captions for subsequent processing.

\textbf{Automatic QA Generation.} 
We prompt GPT-4o to generate multiple (maximum 3 to limit the cost) question-answer pairs pertaining to different aspects of an event caption. We then remove some redundant questions according to the cosine similarities (with a threshold of 0.9) of their CLIP embeddings. Table~\ref{tab:prompt1} shows the prompt for automatic QA generation. After that, to facilitate evaluation, for each generated question, we further curate four additional distractor answers as part of the multiple-choice selection. 
To ensure effectiveness, we incorporate dedicated strategies to obtain the distractor answers:

\begin{table}[t!]
\begin{minipage}{1.0\linewidth}\vspace{0mm}
\centering
\begin{tcolorbox} 
    \small
I need your help in generating question-answer pairs pertaining to the visual event descriptions. Below are the examples: \\
Given description: \{\emph{Event descriptions}\} \\
Good generated Question-Answer (QA) pairs can be: \\
\{\emph{Examples of generated QA pairs}\} \\
Please generate up to 3 QA pairs for each description, and limit the generated questions to a maximum of 22 words while the answers to a maximum of 6 words. 

I hope your questions feature different causal and temporal reasoning keywords such as 'why' and 'how', 'before' and 'after'. Different questions should be diverse and related to different aspects of the described events. Also, make sure the answer is correct according to the description. ... Please label each question in sequence.
Here are the descriptions: \emph{\{descriptions\}}.
\end{tcolorbox}
\caption{Prompt for question generation.}
    \label{tab:prompt1}
\end{minipage}
\vspace{-0.5cm}
\end{table}

For each question, we first obtain 10 candidate answers that are semantically reasonable to answer the question. We achieve this by retrieving the top 10 most similar questions about the same video and borrowing their answers to form the candidate answer list. If there are fewer than 10 QAs of the same video, we fetch the most similar QAs from other videos. 
The question similarity is calculated by the cosine score of their CLIP embeddings.
Then, we pick two candidate answers that are mostly relevant to the video but are not present within the question time span as \textbf{two} distractor answers.
This is achieved by using the segment-irrelevant video content (frames outside the question time span) to retrieve its top-two closest candidate answers (again via cosine similarity of their CLIP embeddings).
Similarly, we pick \textbf{one} candidate answer that is mostly close to the video content specified by the question time span, but is not the same as the correct answer.
Finally, we add \textbf{one} more randomly sampled candidate answer from the whole answer set to ensure that it does not overlap with the existing distractor answers and the correct answer (via simple lexical proximity).

\begin{figure*}
    \centering
    \includegraphics[width=\textwidth]{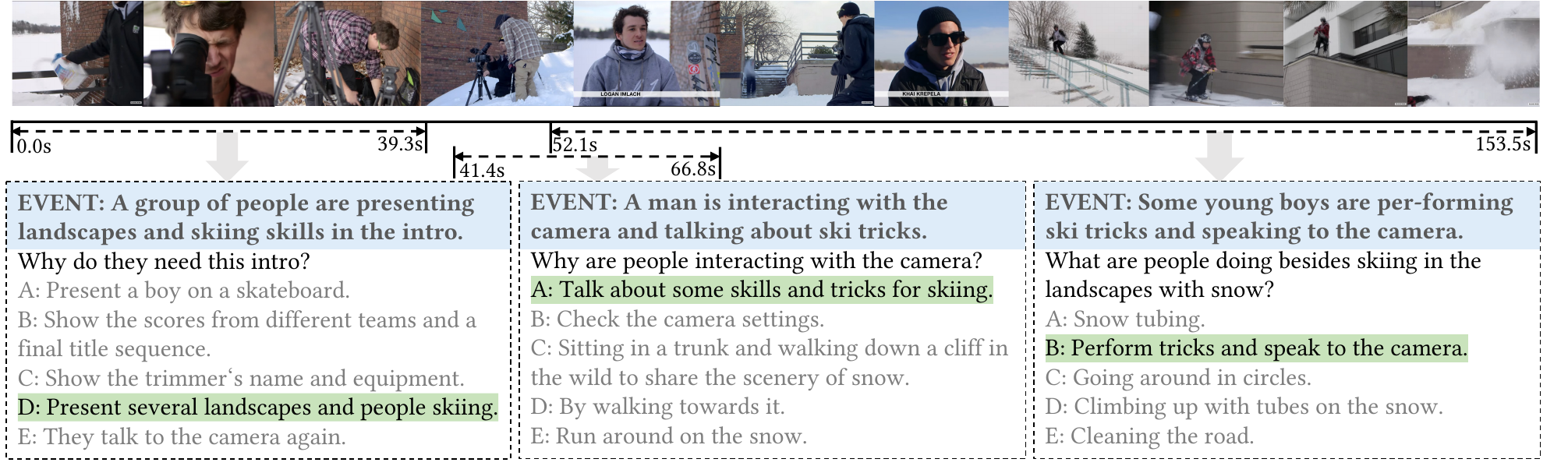} 
    \vspace{-0.6cm}
    \caption{QA examples in \datasetname.}
    \label{fig:newsample}
\end{figure*}

\textbf{Manual QA Checking and Curation.}
As all QAs are automatically generated, we further conduct a manual check and correction to the test QAs. 
Our criteria are to ensure that the distractor answers: 
1) are not potentially correct;  
2) can logically respond to the given question;
3) are distinct from one another; 
4) are closely related to the video content. 
The checking is done by 35 annotators over the course of 267 hours. Around 74\% of QA pairs are modified, with examples of the final QAs shown in Figure \ref{fig:intro} and \ref{fig:newsample}.

\begin{figure}[t]
\centering
\includegraphics[width=\linewidth]{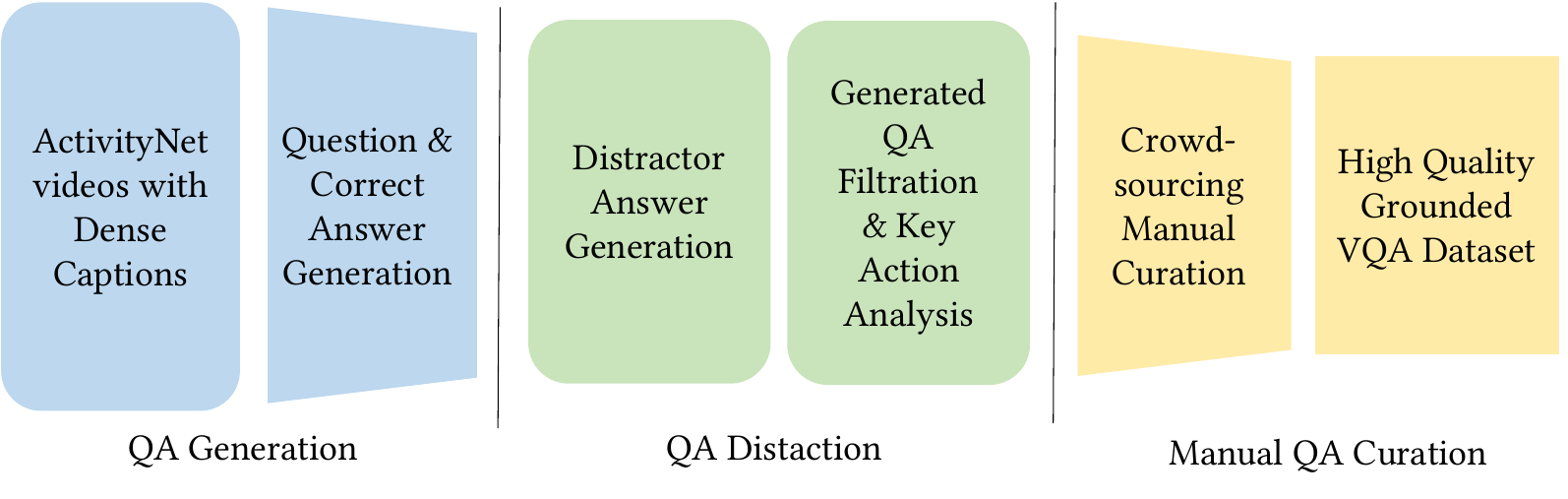}
\caption{\datasetname~construction pipeline.}
\label{fig:data-construct}
\end{figure}

\subsection{Statistics and Analysis}
\begin{table}[t!]
\centering
\begin{adjustbox}{max width=\linewidth}
\begin{tabular}{@{}lccccccc@{}}
\toprule
Split & \# Vid. & \# Que. &
\# Avg. QLen & Seg. Dur.(s) & Vid. Dur.(s) & Ratio (S./V.) \\
\midrule
Train & 7,179 & 53,361 & 10.70 & 38.68 & 127.32 & 0.32 \\
Test & 3,464 & 24,963 & 10.71 & 40.98 & 125.03 & 0.34 \\
\bottomrule
\end{tabular}
\end{adjustbox}
\caption{Statistics of \datasetname. Ratio (S./V.): Average length of segments w.r.t. the entire video.}
\vspace{-0.5cm}
\label{tab:data-stats}
\end{table}

\begin{figure}[t!] 
\centering 
\vspace{-0.35cm} 
\subfigtopskip=0pt 
\subfigbottomskip=0pt 
\subfigcapskip=0pt 
\subfigure[]{
\label{fig:data-stats}
\includegraphics[width=0.43\linewidth]{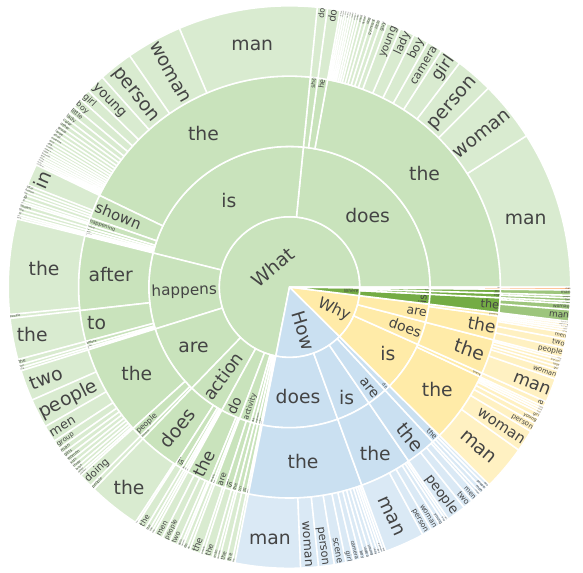}}
\subfigure[]{
\label{fig:certificate}
\includegraphics[width=0.54\linewidth]{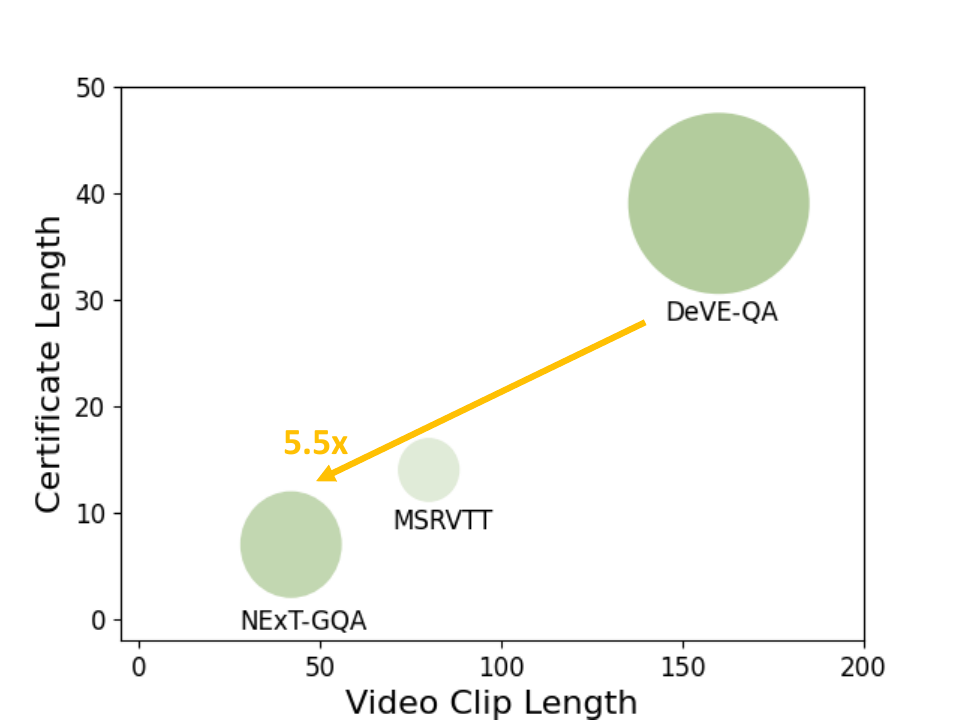}}
\vspace{-0.3cm}
\caption{\datasetname~ analysis. (a) Question distribution in \datasetname. (b) Certificate length of VideoQA datasets. }
\label{level}
\vspace{-0.3cm}
\end{figure}

\datasetname~ is the first 
benchmark dataset that supports question-answering of dense events in long videos.
Table~\ref{tab:data-stats} shows detailed statistics of our \datasetname~ dataset. It comprises 10.6k (7.2k training / 3.5k testing) videos and 78.3k (53.3k training / 25k testing) questions. 
The average video length is 127s and more than 580 videos range from 4-10 minutes. 
The average number of questions per video is 7.5, and the average number of events per video is 2.6 (\vs~ 1 for most other benchmarks). A comparison between these two suggests that an average of 2.5 questions are posed about an individual event.
Figure~\ref{fig:data-stats} shows the distribution of question types; questions are not only about ``what is done'' but also go beyond that to infer ``how" and ``why" questions to target a more comprehensive understanding of events. Note that the ``when" questions are hidden in the requirement on temporal grounding. Also, we limit the number of ``who" and ``where" questions to keep them in a low percentage of the dataset, as they 
can be well-answered without the need for \emph{video}-level understanding~\cite{xu2017video,lei2018tvqa}.

\subsection{Comparison with Existing Benchmarks}\label{sec:EB}
Table~\ref{tab:data-compare} compares \datasetname~with existing VideoQA datasets. 
First and foremost, \datasetname~targets at dense event and long-form VideoQA and enables temporal grounding evaluation. This requirement stands out from all existing datasets that focus on global video events (\eg, all datasets in the 1st block except for NExT-QA) and short videos (\eg, the top-3 datasets listed in Table~\ref{tab:data-compare}). 
Compared with other temporal grounding datasets such as TVQA \cite{lei2018tvqa} and NExT-GQA \cite{xiao2023can}, \datasetname~has longer videos and segments, shaping its challenge for event-level QA. 
For example, Figure~\ref{fig:certificate} shows that the temporal certificate length (average length of video segments needed to answer a question \cite{mangalam2024egoschema}) of \datasetname~is 5.5$\times$ that of NExT-GQA \cite{xiao2023can}. In addition, 
TVQA pays attention to simple visual recognition of ``what is'' in TV shows. Its temporal grounding is biased to localizing the subtitles invoked in the QAs.
\setlength{\tabcolsep}{7pt}
\begin{table}[t!]
\centering
\small
\begin{tabular}{@{}lccccc@{}}
\toprule
Dataset &  D.E. & Vid. Dur.(s) & \#QAs  & Seg. Len(s)  \\ \midrule
MSRVTT-QA~\cite{xu2017video} & \xmark  & 15 &  243K   &  \xmark   \\ 
MSVD-QA~\cite{xu2017video} & \xmark   & 10 & 50K   &  \xmark   \\ 
TGIF-QA~\cite{jang2017tgif}  & \xmark   & 3 & 139K  &  \xmark \\ 
ActivityNet-QA~\cite{yu2019activitynet} & \xmark & 118  &  58K   & \xmark  \\
NExT-QA~\cite{xiao2021next} & \xmark   & 44 & 52K   &  \xmark \\
\midrule
TVQA~\cite{lei2018tvqa}  & \xmark & 76 & 152k  &  11.2 \\ 
NExT-GQA~\cite{xiao2023can}  & \xmark & 42 &  43K   &  7.0  \\ 
\midrule
\datasetname~ (ours) & \cmark & 127 & 78K & 39.4 \\
\bottomrule
\end{tabular}
\caption{Dataset comparison. D.E.: dense event.}
\label{tab:data-compare}
\vspace{-0.8cm}
\end{table}

\section{DeVi Solution} \label{sec:method}
\subsection{Overview}
\begin{figure*}[t!]
\centering
\includegraphics[width=\textwidth]{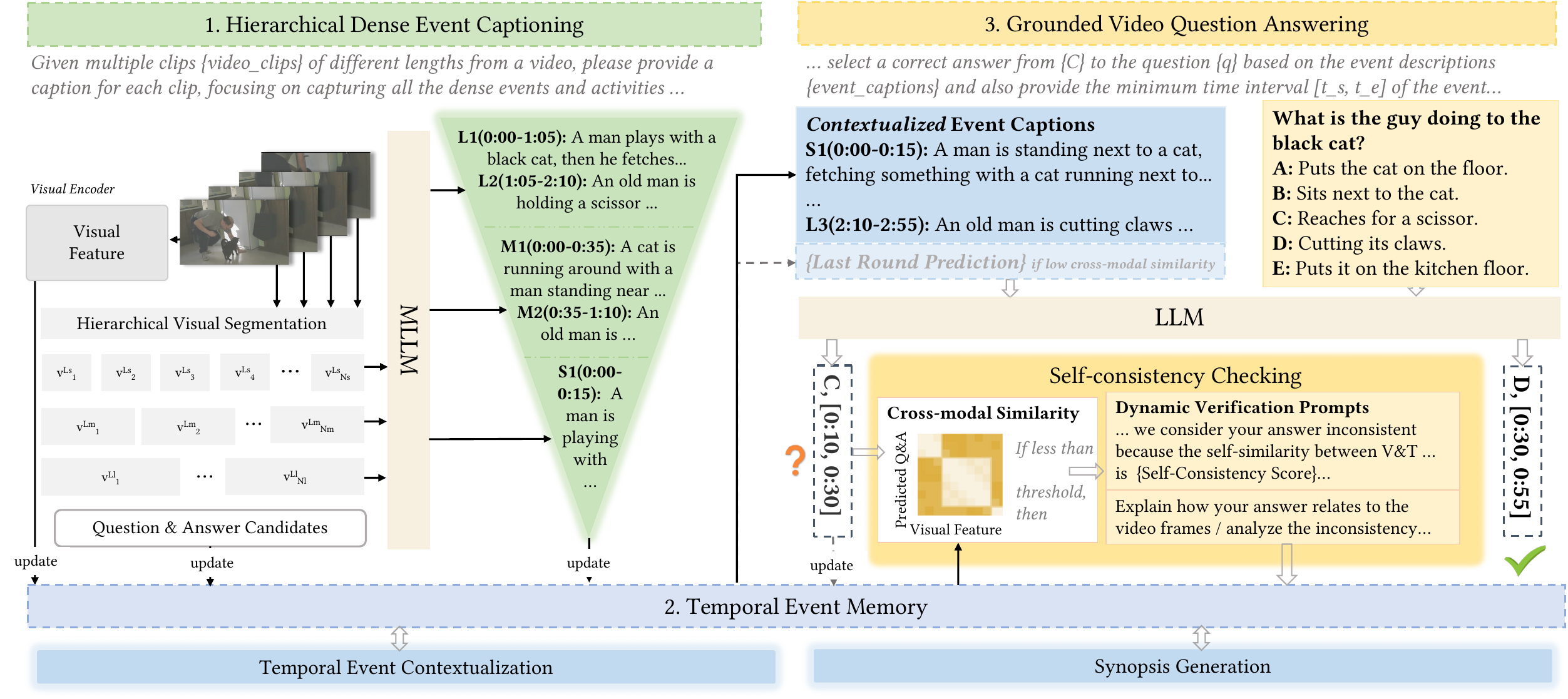} 
\vspace{-0.5cm}
\caption{
\modelname~ framework: (1) Hierarchical dense event video segmenting and captioning, (2) contextualizing and memorizing events in temporal event memory, and (3) event-grounded video question answering with self-consistency checking.}
\label{fig:framework}
\vspace{-0.2cm}
\end{figure*}
Given a $T$-second video $v$ containing a collection of events $E=\{e_1, e_2, \cdots, e_n\}$, a question $q$ along with candidate answer set $C=\{c_1, \cdots, c_5\}$, dense video-event QA predicts the correct answer $\hat{c} \in C$ and grounds it to the relevant event moment
$\hat{t} = \{t_s, t_e\}$ where $t_s \leq t_e \leq T$.  The solution can be formalized as:
\begin{equation}
\label{eqn:obj}
    \{\hat{c},\hat{t}\} = \psi(c,t| E, q, C)\phi(E|v),
\end{equation}
where $\phi$ and $\psi$ denote the models for dense event detection and event-conditioned QA, respectively. Note that the time stamps $t$ come along with the detected events $E$. 

The second term in Eqn.~(\ref{eqn:obj}), $\phi(E|v)$, represents the detection of events $E$ for video $v$. 
To be specific, we incorporate a \textbf{hierarchical dense captioning} (Sec.~\ref{sec:HDES}) mechanism into the MLLM to detect the video events at multiple different time scales. We then design a \textbf{temporal event memory} (Sec.~\ref{sec:TEMM}) module that captures event dependencies (potentially over longer ranges) to contextualize and also memorize the detected video events $E$.
The first term, $\psi(c, t|E,q,C)$, represents 
the \textbf{event-grounded QA} (Sec.~\ref{sec:GQA}) based on the contextualized events $E$, question $q$, and answer set $C$.  
To achieve this, we read from the memory of the 
contextualized events $E$, and feed it to LLMs along with the QAs (question $q$ and candidate answers $C$) to determine the correct answers and the corresponding event moments. In this process, we highlight a self-consistency checking mechanism to ensure the right answer for the right event. An overview of our solution is illustrated in Figure~\ref{fig:framework}. 

\subsection{Hierarchical Dense Event Captioning} \label{sec:HDES}

Dense events within videos are often intertwined and vary in duration. To successfully detect these events, we apply MLLMs (\eg, LLaMA-VID \cite{li2024llamavid}) at multiple scales of temporal hierarchies. 
We build a three-level hierarchy and 
detect events by captioning different lengths of video segments at different hierarchies. 
Our captioning starts from the bottom hierarchy for \emph{short} video segments $V_s = \{v^{L_s}_k\}_{k=1}^{N_s}$.  The short segments 
$V_s$ are given as input to MLLMs along with prompts for describing the segments. 
The corresponding events are denoted as $E_s = \{e^{L_s}_k\}_{k=1}^{N_s}$, where 
$N_s$ and $L_s$ are the number and length of \emph{short} video segments respectively. 
A specific event $e_k$ is given by its text description along with the corresponding start $t_s$ and end $t_e$ time stamps.

We similarly caption \emph{medium} and \emph{long} video segments 
to obtain respective events $E_m = \{e^{L_m}_k\}_{k=1}^{N_m}$ and $E_l = \{e^{L_l}_k\}_{k=1}^{N_l}$, to form a collection of events $E=\{E_s, E_m, E_l\}$ for each video.  Note that $L_s < L_m < L_l \leq T$.
Specifically, our \modelname~partitions the video into non-overlapping segments 
(\eg, 15s/35s/65s for short/medium/long events for \datasetname).  
Then, 5/7/13 frames are sampled uniformly from each segment and sent to LLaMA-VID~\cite{li2024llamavid} for captioning, 
with dedicated prompts to capture different levels of event information depending on the segment length. 

\subsection{Temporal Event Memory} \label{sec:TEMM}
The events so far have been independently detected by focusing on individual video segments.
The lack of contextual information may result in inaccurate or incomplete event captions. While the hierarchical captioning strategy helps alleviate the issue, it cannot model long-term temporal event dependencies.
For example, the video in Figure~\ref{fig:intro} may result in the event of ``a man enters the field" at the beginning and ``a biker is performing" in the middle of the video. However, questions about why the man enters the field and who (man or woman) the biker cannot be determined based on the individual event captions without linking the two events.

To capture long-term dependencies, we design a memory module to contextualize the event captions while storing the visual and event representations. 
The LLM is asked to refine the captions by prompting ``... given a set of event captions \{E\} and a question \{q\} of a video, you are required to refine each caption by incorporating contextual information from all the other captions and question via analyzing the overall narratives, identifying relevant context and incorporate context with coherence...". We also curate examples to perform in-context learning for the LLMs before the actual generation.
We additionally prompt GPT-4o to articulate all the events into a synopsis $e_y$ to serve as a global event for the entire video. Our specific prompts are given in Table \ref{tab:prompt5}. Consequently, we obtain $E' = \{E_s^{'}, E_m^{'}, E_l^{'}, e_y\}$, in which the events at each level are enhanced with long-range temporal dependency. 

To be specific, the hierarchical video event captions $\{E\}$ are initially updated to the temporal event memory. 
At the same time, we sample the original video at 1 fps and encode the frames into visual representations with CLIP VIT-L/14~\cite{radford2021learning}.  The representations are denoted as $f_v$ and are stored in the temporal event memory.  They will be read by the subsequent self-consistency checking module (see Sec.~\ref{sec:GQA}) to help determine the correct answers according to the cross-modal similarity between answers and visual representations. Overall, the temporal event memory caches different representations (visual features, dense captions, and global synopsis) $M = \{E,E^{'},f_v\}$ from videos to aid in answering questions and grounding results about specific events.

\begin{table*}[t!]
\begin{minipage}{1.0\linewidth}\vspace{0mm}
\centering
\begin{tcolorbox} 
    \small
You are a highly intelligent language agent in improving the quality of video captions. 
Given a set of captions (each representing a different time segment of a video) and a question about the video, you are required to refine each caption by incorporating contextual information from all the other captions and questions via analyzing the overall narrative, identifying relevant context and incorporating context with coherence. Here are the captions and questions: \$\{hierarchical captions\} and \{question\}. Here are the examples:
\begin{itemize}
\item Original Caption: A person is holding a knife and waving it around.\\
Contextualized Caption: A person is holding a knife and chopping down a tree.
\item Original Caption: A person takes off their clothes by the river and jumps into the water to swim.\\
Contextualized Caption: A person takes off their clothes by the river and jumps into the water to save someone who is drowning.
\item Original Caption: A person is waving a spatula in the kitchen.\\
Contextualized Caption: A person is using a spatula in the kitchen to chase away a squirrel that has entered.
\end{itemize}

Please provide a comprehensive synopsis of the entire video based on the captions, covering all key temporal actions, characters, and interactions.
\end{tcolorbox}
\vspace{-2mm}
\caption{Prompt of temporal contextualization.}
    \label{tab:prompt5}
\end{minipage}
\vspace{-0.7cm}
\end{table*}

\subsection{Event-Grounded QA and Consistency Check} \label{sec:GQA}
Intuitively, the events $E'$ can be read from the event memory and fed into an LLM (\eg, GTP-4o or Gemini 2.0)  
along with the QAs to determine the answers and their supportive visual evidence. 
For example, this can be achieved via the prompts 
``... select a correct answer from the candidate answer set $\{C\}$ to the question $\{q\}$ based on the event set $\{E\}$ and also output the time span $[t_s, t_e]$ of the event that carries the correct answer ...".
While straightforward,  we find that this approach does not perform well; there is a large discrepancy where the LLM often gives the correct answer but with wrong grounding or vice versa.  
We therefore perform consistency checking between a predicted answer and the time span. 

Specifically, we evaluate consistency based on the cosine similarity $R_{va}$ between the answer $a$ and the video content within the predicted time span $[t_s, t_e]$:
$
    R_{va} = cos(f_v, f_a) = \frac{f_v \cdot f_a}{||f_v||||f_a||},
$
where $f_a$ and $f_v$ are encodings of the answer text and video segment using CLIP~\cite{radford2021learning}.
Predictions with low consistency (\ie, small $R_{va}$) are sent back to the LLM for adjustment. 
This process is iterated until the consistency reaches a threshold $\sigma$ or a maximum number of iterations $\delta$.   
While $R_{va} < \sigma$, we resubmit the captions and QA pair with a \emph{dynamic verification prompt} (Table~\ref{tab:prompt4}) to the LLM. 
The verification prompt is to exploit the self-consistency results from the previous round for better answer reasoning. 
The model is asked to re-answer the question with the extra information. 
If the results from the two rounds are consistent, the model needs to elaborate on the relationship between the answer and the video segments. Otherwise, it is asked to explain the inconsistency.

\setlength{\tabcolsep}{1pt}
\begin{table}[t!]
\centering
\begin{adjustbox}{max width=\linewidth}
\begin{tabular}{@{}lccc@{}}
\toprule
Model & Frames & LLM & Acc@QA \\
\midrule
\textcolor{gray}{\textit{Open-source}} & & & \\
VideoLLaMA~\cite{zhang2023video}& 8 & Vicuna-7B & 41.2 \\ 
InternVideo~\cite{wang2023internvid} & 16 & CLIP Text Encoder & 48.3 \\
VFC~\cite{momeni2023verbs} & 32 & PaLM  & 49.5 \\
Video-LLaVA~\cite{lin2023video} & 8 &Vicuna-7B  & 56.2 \\
LLaVA-NeXT-Video~\cite{liu2024llava} & 32 & Vicuna-7B  & 57.1 \\
LLaMA-adapter (SFT)~\cite{zhang2023llama}  & - & LLaMA-7B & 58.3 \\
Videochat2~\cite{li2024mvbench} & 16 & Vicuna-7B & 58.7 \\
SeViLA~\cite{yu2024self} & 32 & BLIP-2 & 61.2 \\
VideoLLaMA2~\cite{cheng2024videollama} & 16 & Mistral-7B-Instruct & 61.3 \\
LLaVA-OV ~\cite{li2024llava} & 16 & Qwen2-7B & 61.9 \\
Qwen2-VL~\cite{wang2024qwen2} & 32 & Qwen2-7B & 63.5\\
PLLaVA~\cite{xu2024pllava} & 16 & LLaVA-Next-7B & 63.7 \\
LongVA~\cite{zhang2024long} & 48 & Qwen2-Extended &64.9 \\
\midrule
\textcolor{gray}{\textit{API-based}} & & & \\
ViperGPT~\cite{suris2023vipergpt} & - & GPT-3 & 55.1  \\
IG-VLM~\cite{kim2024image} & 6 & LLaVA-1.6-7B & 60.2\\
GPT-4o~\cite{GPT-4o} & 16 & GPT-4o & 62.6 \\
Gemini-2.0~\cite{gemini2} & - & Gemini-2.0 & 63.4\\
LLoVi~\cite{zhang2023simple} & - & GPT-4  & 63.8  \\
VideoAgent~\cite{wang2025videoagent} & - & GPT-4 & 64.5 \\
\rowcolor{lightgray}
\modelname-GPT-4o (ours) & - & GPT-4o & 71.2\\
\rowcolor{lightgray}
\textbf{\modelname-Gemini-2.0 (ours)} & - & Gemini-2.0 & \textbf{71.8}\\
\bottomrule
\end{tabular}
\end{adjustbox}
\caption{QA Results on \datasetname.}
\label{tab:experi-aqa}
\vspace{-0.82cm}
\end{table}

\begin{table*}[t!]
\begin{minipage}{1.0\linewidth}\vspace{0mm}
\begin{tcolorbox} 
    \small
You are a helpful expert in dense event video analysis. You have been provided with video descriptions and one multiple-choice question about the video and gave out your answer and the minimum frame(s) interval to support. However, after our professional check, we consider your answer inconsistent because the self-similarity between your previous answer \{\emph{Previous\_Answer}\} and \{\emph{Supportive\_Frames}\} is only \{\emph{Self\_Consistency\_Score}\}.

On this premise, I want you to answer this question again: \{\emph{Prompts\_for\_Event-Grounded\_QA}\} and judge whether your answer is consistent with the previous one. If no, analyze the inconsistency in detail. If yes, explain how the answer relates to the video frames.
\end{tcolorbox}
\vspace{-2mm}
\caption{ Prompt for dynamic verification.}
    \label{tab:prompt4}
\end{minipage}
\vspace{-0.5cm}
\end{table*}

\vspace{-0.3cm}
\section{Experiments}

\subsection{Configuration and Evaluation}
We experiment on the test set of \datasetname. Additionally, we extend our experiments to 
NExT-GQA~\cite{xiao2023can}. NExT-GQA features temporally grounded QA about multiple actions. It contains 990 videos and 5,553 questions for testing.
For hierarchical event captioning, the segment lengths $L_s$, $L_m$ and $L_h$ are set to \{10s, 35s, 65s\} for \datasetname~and \{5s, 15s, 45s\} for NExT-GQA. 
For self-consistency checking, $\sigma$ is empirically set to 0.4 (refer to Figure~\ref{fig:group}(d)), and $\delta$ is set to 2 for efficiency. 
For evaluation, we follow NExT-GQA~\cite{xiao2023can} to report QA accuracy Acc@QA, grounding quality Intersection over Prediction (IoP) and Union (IoU), as well as grounded QA accuracy Acc@GQA, all in percentages (\%). In particular, Acc@GQA stands for the percentage of questions that are correctly answered and visually grounded, \ie, the predicted time span overlaps the ground-truth event moment with an IoP $\geq$ 0.5. 

\subsection{Performance Analysis}

\setlength{\tabcolsep}{2pt}
\begin{table}[t!]
\centering
\begin{adjustbox}{max width=\linewidth}
\begin{tabular}{@{}lcccccc@{}}
\toprule
Model &  mIoP & IoP@0.5 & mIoU & IoU@0.5 & Acc@QA & Acc@GQA \\
\midrule
Human & 58.2 & 62.3 & 43.9 & 52.7 & 84.7 & 62.4\\
\midrule
\textcolor{gray}{\textit{Weakly-supervised}} &  &  & \\
FrozenBiLM(NG+)~\cite{xiao2023can} & 21.2 & 18.2 & 8.50 & 6.2 & 61.6 &14.5 \\
Temp[CLIP](NG+)~\cite{xiao2023can} & 24.6 & 24.8 &12.5 & 9.1& 58.9 &14.9 \\ 
SeViLA*~\cite{yu2024self} & 25.8 & 19.9 & 21.2 & 11.5 & 62.7 & 16.1 \\
\midrule
\textcolor{gray}{\textit{Zero-shot}} &  &  &  \\
LLaVA-Next-Video~\cite{liu2024llava} & 22.5 & 21.1 & 13.8 & 10.7 & 56.9 & 17.4\\
VideoChat2 & 23.1 & 21.8 & 14.2 & 12.5 & 59.2 & 18.6\\
VideoLLaMA2~\cite{cheng2024videollama}  & 23.7 & 22.0 & 12.9 & 10.1 & 62.0 & 19.2\\
Qwen2-VL~\cite{wang2024qwen2} & 23.6 & 23.2 & 16.5  & 14.7 & 63.9 & 20.1\\
LongVA~\cite{zhang2024long} & 24.9 & 24.7 & 16.9 & 15.2 & 66.2 & 20.8\\
LLoVi~\cite{zhang2023simple} & 27.5 & 27.0 & 17.9 & 13.0 & 63.9 & 22.9\\
\rowcolor{lightgray}
\modelname-GPT-4o (ours) &33.8& 32.2& 20.7 &17.4 &71.9 & 27.1 \\
\rowcolor{lightgray}
\textbf{\modelname-Gemini-2.0 (ours)}  & \textbf{34.9} & \textbf{32.8} & \textbf{21.7} & \textbf{18.5} & \textbf{72.1} & \textbf{27.7} \\
\bottomrule
\end{tabular}
\end{adjustbox}
\caption{Grounded VideoQA results on \datasetname. *: pre-trained on video grounding datasets. Human results are obtained on a subset of random 3K questions.}
\label{tab:experi-agqa}
\vspace{-1.5cm}
\end{table}

We adapt prominent MLLMs (\eg, Video-LLaMA2 \cite{cheng2024videollama}, LLaVA-NeXT-Video \cite{zhang2024llavanextvideo}, GPT-4o \cite{GPT-4o}, LongVA~\cite{zhang2024long}, and etc) that perform well on ``single event" QA to \datasetname~and compare them with \modelname. The models (except for LLaMA-Adapter \cite{zhang2023llama}) are directly prompted for zero-shot VideoQA with the instruction like ``... select a correct answer from \{C\} to the question \{q\} and provide the minimum time interval based on the visual content...". LLaMA-Adapter is finetuned on \datasetname~training set to serve as a reference result of the finetuned model. In our implementation, we uniformly sample 16 frames and aggregate their representations via mean-pooling to adapt to videos. 
GPT-4o also takes 16 frames for question answering.
For Gemini 2.0, the raw videos are directly input for question answering. For other methods, we follow their official optimal protocols for video input but within our computational affordance.

As most of these methods cannot perform grounding, we compare them with QA accuracy alone.
Table~\ref{tab:experi-aqa} shows that \modelname, with an accuracy of 71.8\%, outperforms the second-best model LongVA~\cite{zhang2024long} (designed for long video understanding) significantly by 6.9\%. Moreover, \modelname~surpasses it's two closest baselines VideoAgent \cite{wang2025videoagent} and LLoVi~\cite{zhang2023simple} remarkably by 7.3\% and 8.0\% respectively. We also find that all other end-to-end MLLMs such as Video-LLaMA, Video-LLaVA, and VideoChat2 perform much worse than \modelname~by 10\% $\sim$ 30\%. The results demonstrate that \modelname~has made significant optimizations over general MLLMs in performing dense-event question-answering on long-range videos. 

Table~\ref{tab:experi-agqa} presents grounded QA accuracy, compared with methods from \cite{xiao2023can}.
\modelname~surpasses SoTA zero-shot method LLoVi by 4.8\% on Acc@GQA. 
Improvements come from both better QA (+8.9\% Acc@QA) \emph{\textbf{and}} better grounding (+5.8\% IoP@0.5). This differs from the previous methods, where improvements are primarily from either better grounding \emph{\textbf{or}} better QA alone (also see Table \ref{tab:experi-nextgqa} on NExT-GQA). Interestingly, we find that the zero-shot methods generally outperform the weakly-supervised methods. This indicates that the weakly-supervised methods are highly likely to learn shortcuts from questions to answers. It also reflects the power of LLMs in question answering. 
Finally, we find that there is a clear performance gap between existing methods and humans, especially in performing grounded QA (up to 34.7\%). This indicates the severe inadequacy of existing algorithms in reliable dense-event QA.


We also extend a comparison on NExT-GQA. Table~\ref{tab:experi-nextgqa} shows that \modelname~consistently outperforms other competitors on NExT-GQA, even though its superiority gets shrunk a little bit compared to that on \datasetname~(\eg, DeVi surpasses LLoVi by 4.8\% on DeVE-QA and 2.1\% on NExT-GQA regarding Acc@GQA). 
The results demonstrate \modelname's effectiveness and strength in fine-grained action reasoning in short videos aside from dense-event long video understanding.

\subsection{Ablation Studies}
\label{sec:AS}

\begin{table}[t!]
\centering
\small
\begin{adjustbox}{max width=\linewidth}
\begin{tabular}{@{}lcccccc@{}}
\toprule
Model &  mIoP & IoP@0.5 & mIoU & IoU@0.5 & Acc@QA & Acc@GQA \\
\midrule
Human & 72.1 & 86.2 & 61.2 & 70.3 & 93.0 & 82.0\\
\midrule
\textcolor{gray}{\textit{Weakly-supervised}} &  &  & \\
IGV \cite{li2022invariant} & 21.4 & 18.9 & 14.0 & 9.6 & 50.1 & 10.2 \\
VGT \cite{xiao2022video} & 25.3 & 25.3 & 3.0 & 1.7 & 55.7 & 14.4 \\
Temp[CLIP](NG+)~\cite{xiao2023can} & 25.7 & 25.5 & 12.6 & 8.9 & 60.2 & 15.9 \\ 
SeViLA*~\cite{yu2024self} & 29.5 & 22.9 &21.7 & 13.8 & 68.1 & 16.6 \\
FrozenBiLM(NG+)~\cite{xiao2023can} & 24.2 & 23.7 & 9.5 & 6.1 & 70.8 & 17.5 \\
QGAC-TR \cite{xu2024exploring} & 28.3 & 27.7 & 15.7  & 11.7 & 63.6 & 18.3 \\ 
FrozenBiLM(TimeCraft)~\cite{liu2025timecraft} & 26.3 & 24.9 & 13.2 & 8.4 & \textbf{74.7} & 18.5 \\
\midrule
\textcolor{gray}{\textit{Zero-shot}} &  &  &  & \\
VideoStreaming~\cite{qian2024streaming} & 32.2 & 31.0 & 19.3 & 13.3 & - & 17.8\\
LLoVi~\cite{zhang2023simple} & 39.4 & 38.0 & 21.5 & 16.2 & 73.8 & 26.8 \\
\rowcolor{lightgray}
\textbf{\modelname-GPT-4o (ours)} & 39.3 & 37.9 & 22.3 & 17.4 & 71.6 &28.0\\
\rowcolor{lightgray}
\textbf{\modelname-Gemini-2.0 (ours)} & \textbf{39.7} & \textbf{38.9} & \textbf{23.6} & \textbf{19.5} & 73.1 & \textbf{28.9}\\
\bottomrule
\end{tabular}
\end{adjustbox}
\caption{Grounded VideoQA results on NExT-GQA.}
\label{tab:experi-nextgqa}
\vspace{-0.6cm}
\end{table}

\setlength{\tabcolsep}{8pt}
\begin{table}[t!]
\centering
\small
\begin{adjustbox}{max width=\linewidth}
\begin{tabular}{@{}lccc@{}}
\toprule
Model Variants &  Acc@QA  & Acc@GQA \\
\midrule
\modelname~  & \textbf{71.8} & \textbf{27.7}  \\ 
\hspace {0.5cm}  w/o Hierarchical Dense Captioning  & 66.9 & 23.3 \\
\hspace {0.5cm}  w/o Temporal Contextualizing  & 68.8 & 25.3 \\
\hspace {0.5cm} w/o Consistency Checking & 66.3 & 21.7 \\
\bottomrule
\end{tabular}
\end{adjustbox}
\caption{ Major model ablation on \datasetname.}
\label{tab:major_aba}
\vspace{-0.8cm}
\end{table}

Table \ref{tab:major_aba} presents the ablation results of the three major components in \modelname~on \datasetname.
Substituting the \textbf{hierarchical dense captioning} with a strategy of naively captioning every frame (as in LLoVi \cite{zhang2023simple}) decreases QA and GQA accuracy by 4.9\% and 4.4\%, respectively. Further in-depth analyses in Table~\ref{tab:qtype} verify that hierarchical dense captioning facilities answering questions about denser event videos. Without it, the QA performance drops by 5.7\% on the video subset with denser events, 
whereas the performance only drops by 2.3\% and 3.7\% on the video subsets with sparse events (1 to 2 events).
This further demonstrates its ability to capture specific information from multiple and complicated events.

Removing the \textbf{temporal event contextualization} module degrades results by 3.0\% on QA and 2.4\% on GQA. This is understandable as the contextualized captions are with less misunderstanding and incompletion problems compared with isolated captionings. 
Moreover, the ablation results (\eg, -3.7\% on long video GQA \vs -1.1\% on short video GQA) in Table~\ref{tab:vlength} also justify its effectiveness in handling long videos.
Finally, we remove the \textbf{self-consistency checking} module 
and apply a naive way to prompt LLMs for final predictions. The QA and GQA accuracy degenerate significantly by 5.5\% and 6.0\% respectively, suggesting that a large number of LMM's answers are not anchored on the relevant video content without self-consistency checking.

\begin{figure*}[t!]
\centering
\includegraphics[width=\linewidth]{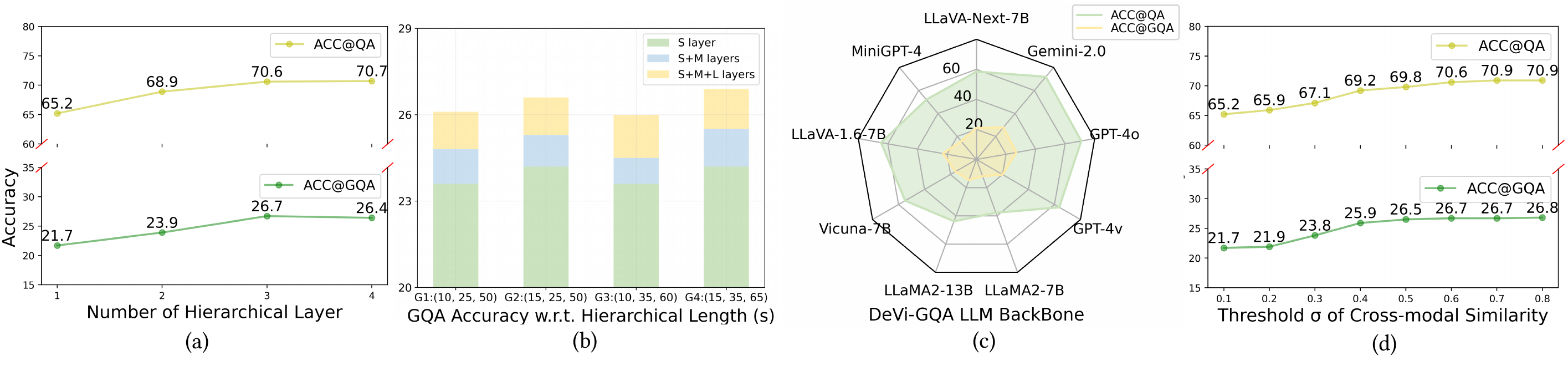} 
\caption{Analysis of \modelname.
(a) Analysis of the number of hierarchy layers.
(b) Analysis of the segment length.
(c) Analysis of MLLM reasoning backbone on \datasetname. 
(d) QA and GQA accuracy w.r.t. cross-modal similarity threshold $\sigma$. }
\label{fig:group}
\end{figure*}

\setlength{\tabcolsep}{7pt}
\begin{table}[t]
\small
\centering
\begin{adjustbox}{max width=\linewidth}
\begin{tabular}{@{}c|c|cccc}
\toprule
\multirow{2}{*}{Metrics} &\multirow{2}{*}{Model} & \multicolumn{4}{c}{Event Density} \\
\cline{3-6}
~ & ~ & Single & Double & Dense & Total  \\
\midrule
\multirow{5}{*}{Acc@QA}
     & FrozenBiLM(NG+)~\cite{yang2022zero} & 62.1 & 61.8 & 59.2 & 61.6 \\
  ~  & SeViLA~\cite{yu2024self} & 63.3 & 62.9 & 61.7 & 62.7 \\
   ~ & LLoVi~\cite{zhang2023simple} & 65.2 & 65.8 & 61.2 & 63.9 \\
  ~  & DeVi w/o HDC & 66.2 & 65.5 & 67.1 & 66.9  \\
  ~  & {\bf DeVi} & {\bf68.2} & {\bf69.7} & {\bf72.8} & {\bf71.8} \\
  \midrule
\multirow{5}{*}{Acc@GQA}
   ~ & FrozenBiLM(NG+)~\cite{yang2022zero} & 15.1 & 15.0 & 13.9 & 14.5 \\
   ~ & SeViLA~\cite{yu2024self} & 15.9 & 16.1 & 16.2 & 16.1 \\
   ~ & LLoVi~\cite{zhang2023simple} & 24.1 & 22.6 & 21.1 & 22.8 \\
   ~ & DeVi w/o HDC & 23.5 & 23.3 & 24.2  & 23.3 \\
   ~ & {\bf DeVi} & {\bf25.8} & {\bf27.0} & {\bf 28.8} & {\bf27.7} \\
\bottomrule
\end{tabular}
\end{adjustbox}
\caption{Results w.r.t. different event densities. Single/Double/Dense-Event: 1/2/more than 2 main event(s) is/are present in the related videos. 200 videos are selected from each subset. HDC: Hierarchical dense captioning.}
\label{tab:qtype}
\vspace{-0.5cm}
\end{table}

\setlength{\tabcolsep}{7pt}
\begin{table}[t]
\small
\centering
\begin{adjustbox}{max width=\linewidth}
\begin{tabular}{@{}c|c|cccc}
\toprule
\multirow{2}{*}{Metrics} &\multirow{2}{*}{Model} & \multicolumn{4}{c}{Video Length} \\
\cline{3-6}
~ & ~ & Short & Medium & Long & Total  \\
\midrule
\multirow{4}{*}{Acc@QA}
     & SeViLA~\cite{yu2024self} & 64.2 & 62.4 & 60.6 & 62.7 \\
   ~ & LLoVi~\cite{zhang2023simple} & 66.0  & 64.1 & 62.8 & 63.9 \\
  ~  & DeVi w/o TC & 68.9 & 68.8 & 68.8 & 68.8\\
  ~  & {\bf DeVi} & {\bf 70.1} & {\bf 71.2} & {\bf72.7} & {\bf71.8} \\
  \midrule
\multirow{4}{*}{Acc@GQA}
   ~ & SeViLA~\cite{yu2024self}  & 18.4 & 16.2 & 14.9  & 16.1 \\
   ~ & LLoVi~\cite{zhang2023simple} & 24.7 & 22.4 & 21.1 & 22.8 \\
   ~ & DeVi w/o TC & 25.4 & 25.5 & 25.2 & 25.3 \\
   ~ & {\bf DeVi} & {\bf25.5} & {\bf27.3} & {\bf 28.9} & {\bf27.7}\\
\bottomrule
\end{tabular}
\end{adjustbox}
\caption{Results w.r.t. different video lengths. Short/Medium/Long: videos that are 0-60/60-120/more than 120 seconds. 200 videos are selected for each event-density level. TC: Temporal Contextualizing.}
\label{tab:vlength}
\vspace{-1cm}
\end{table}

To better dissect the models' behavior in answering questions with different event densities and video lengths, we conduct additional evaluations on video subsets with different event numbers and lengths in Table \ref{tab:qtype} and \ref{tab:vlength}.
The results demonstrate 
\modelname~ is especially strong for 
longer videos with denser events compared to existing methods. 
Specifically,
Table \ref{tab:qtype} shows that the accuracy of existing MLLMs decreases with the increase of event density (\eg, FrozenBiLM(NG+) from 62.1\% to 59.2\%), whereas \modelname's accuracy increases from 68.2\% to 71.8\%. 
Table~\ref{tab:vlength} analyzes performance with different lengths of videos, and shows that \modelname~ maintains a strong performance on medium (60-120 seconds) and long (more than 120 seconds ) videos, while other baselines decrease visibly. This unequivocally shows \modelname's proficiency in handling long videos. 
Additionally, the results also highlight the importance of \modelname's specially designed modules in handling dense events and long videos.

\begin{figure*}[ht!]
\centering
\includegraphics[width=\linewidth]{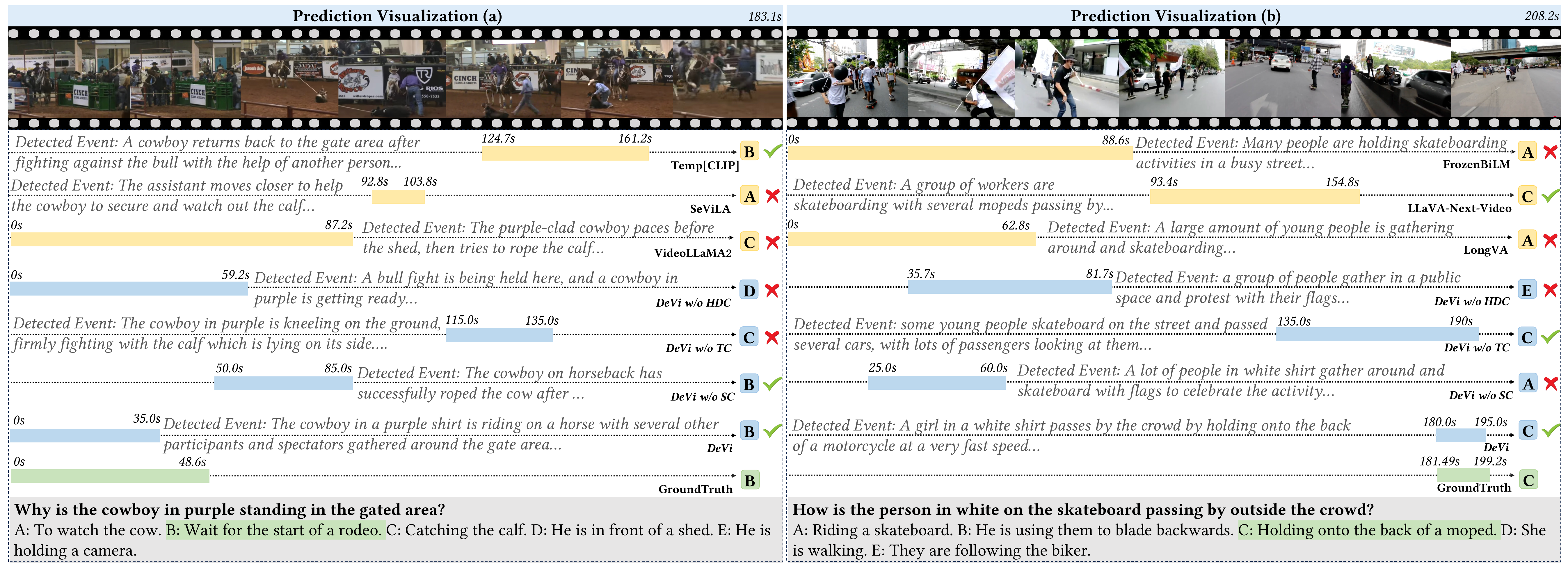} 
\caption{Prediction visualization on \datasetname. 
Baseline models like SeViLA and Temp[CLIP] tend to answer the question without truly grounding it to the relevant video segments. Hierarchical Dense Captioning (HDC) helps \modelname~ understand the events in different scales,  Temporal Contextualizing (TC) helps improve GQA by refining or correcting the isolated captions according to related context, and Self-consistency checking (SC) is effective in correcting wrongly grounded segments.
}
\label{fig:analysis}
\vspace{-0.1cm}
\end{figure*}
The prediction visualization of QA and grounding results between different models, as shown in Figure~\ref{fig:analysis}, demonstrates the efficacy of \modelname, as well as our design with self-consistency checking (in temporal grounding) and hierarchical dense captioning (in dense event QA). 
The self-consistency checking mechanism effectively identifies and corrects wrongly-grounded segments (Figure~\ref{fig:analysis}(a) row 6). At the same time, removing hierarchical dense captioning leads to the summarization of wrong events 
from the video content (Figure~\ref{fig:analysis}(b) row 4). Furthermore, temporal contextualization 
improve both QA accuracy and the grounding of visual information to the relevant segments of the video. 

Qualitative analyses of other baseline models also highlight their limitations. 
For instance, even though VideoLLaMA2~\cite{cheng2024videollama} detected the roughly correct time span of the target event (Figure~\ref{fig:analysis}(a) row 3), its ineffective processing of temporal dynamics leads to its false understanding of the key event content and finally results in a wrong answer. LongVA~\cite{zhang2024long} is good at handling long videos, yet it falls short in coping with obfuscating content with similar characteristics of objects/persons and recognizing the correct event span for other confusing events (Figure~\ref{fig:analysis}(b) row 3).

\subsection{Implementation Investigations}
\begin{table}[t!]
\centering
\small
\begin{adjustbox}{max width=\linewidth}
\begin{tabular}{@{}lccc@{}}
\toprule
Caption Model &  Acc@QA  & Acc@GQA \\
\midrule
VideoBLIP~\cite{yu2023efficient}   & 60.1 & 20.0 \\ 
VideoBLIP~\cite{yu2023efficient} w HDC  & 64.2 & 23.9 \\ 
\midrule
Video-LLaVA~\cite{lin2023video} & 65.8 & 24.1 \\
Video-LLaVA~\cite{lin2023video} w HDC & 70.8 & 26.9 \\
\midrule
LLaMA-VID~\cite{li2024llamavid} & 66.9 & 23.3 \\
LLaMA-VID~\cite{li2024llamavid} w HDC & \textbf{71.8}  & \textbf{27.7} \\
\bottomrule
\end{tabular}
\end{adjustbox}
\caption{Investigation of captioners.}
\label{tab:experi-cap}
\vspace{-1cm}
\end{table}

\textbf{Dense Video Event Captioner} 
Table \ref{tab:experi-cap} shows that substitution of LLaMA-VID \cite{li2024llamavid} with VideoBLIP \cite{yu2023efficient} as \modelname's captioner deteriorates the accuracy by over 4\% and 7\% for QA with and without grounding respectively. We attribute the result of LLaMA-VID being a better captioner to the fact that it is specially trained on long videos. In addition, LLaMA-VID represents frames with context tokens and content tokens. This could benefit to squeeze critical information and thus bring more accurate captions.

We further analyze the influence of hierarchy level and segmentation length on \datasetname. As depicted in Figure~\ref{fig:group}(a), the results peak at 3 hierarchy layers; the hyperparameters are finalized to be 15s, 35s, and 65s with experiments. Additionally, we observe from Figure ~\ref{fig:group}(b) that increasing segment length brings better GQA accuracy (G2 \& G3), indicating that it is influenced by the nature of datasets (overall duration, timestamps, etc.).

\textbf{LLM Backbone} 
We additionally investigate the behavior of \modelname~with different MLLMs to perform QA as described in Sec.~\ref{sec:GQA}. 
Figure~\ref{fig:group}(c) shows
that Gemini achieves the best performance (71.8\% for QA and 27.7\% for GQA), followed by GPT-4o (71.1\%) and GPT-4V (64.2\%). These results again suggest that stronger LLMs (\eg, GPT-4o/4v and Gemini) are key to success, as indicated by the remarkable performance gap in both GQA and QA accuracy between GPT-4/Gemini and other smaller open-source models. 
We also observe that the GQA accuracy improves when increasing the LLM size of the same model (\eg, from 10.9\% of LLaMA2-7B to 14.6\% of LLaMA2-13B). The findings reinforce our belief that larger LLMs have higher zero-shot QA reasoning performance and are vital to the success of \modelname~in dense event reasoning.
\vspace{-0.3cm}
\subsection{Efficiency Analysis}
\begin{figure}[t]
\centering
\scalebox{0.8}{
\includegraphics[width=1.0\linewidth]{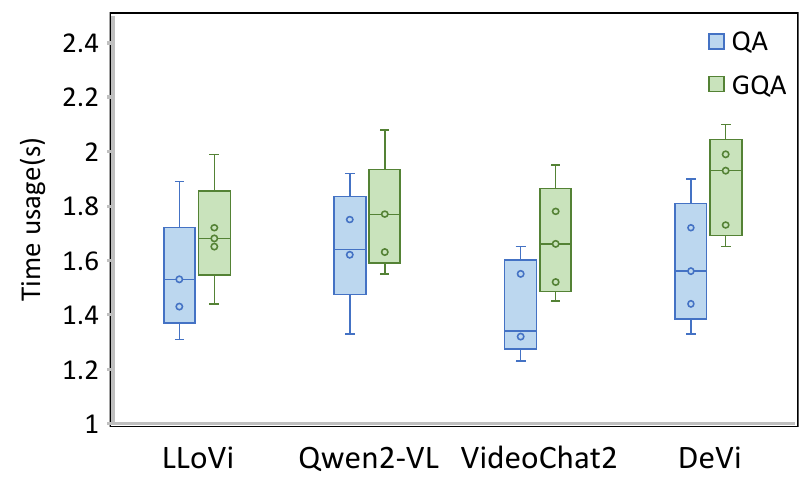}
}
\vspace{-0.3cm}
\caption{Analysis of inference efficiency. All models behave roughly the same, while \modelname~is slightly slower on GQA.}
\vspace{-1.0cm}
\label{fig:timeuse}
\end{figure}

To analyze the efficiency of \modelname, we conduct experiments on the NVIDIA A800 GPU by randomly selecting $1K$ samples. We compare its average inference speed with 3 baseline models. Figure~\ref{fig:timeuse} shows that considering QA efficiency, \modelname~and LLoVi are roughly at the same running time of 1.59s and 1.54s respectively, while Qwen2-VL is the slowest at 1.65s and VideoChat2 is the fastest at 1.41s. Considering GQA where more complex reasoning is needed, all four models take slightly longer due to the slower responses of LLMs, especially for \modelname~ which additionally undergoes self-consistency checking. Further analysis of \modelname reveals that this event-grounded QA and self-consistency process accounts for nearly half of the total runtime. This also reflects that 
our hierarchical dense captioning strategy does not slow down but accelerates over naively captioning every frame as is done in LLoVi, since \modelname~and LLoVi are roughly at the same running speed.

\section{Conclusion}
In this paper, we have introduced a novel task of question answering on dense video events. The task challenges MLLMs from three aspects of dense-event captioning, long-form video understanding, and faithful multimodal reasoning by grounding. To facilitate the study, we construct the \datasetname~dataset with manual efforts and benchmark many advanced MLLMs and show their weaknesses on the proposed task. For improvements, we propose \modelname~, a training-free and modular MLLM approach that solves the aforementioned challenges by a set of tailored practices, including hierarchical dense event captioning, temporal event contextualizing and memorizing, and trustworthy QA with self-consistency checking. Our extensive experiments have demonstrated the effectiveness and superiority of \modelname. Moreover, we share our implementation alternatives and highlight the power of larger LLMs for our success. With these efforts, we hope this work provides a solid foundation for QA research on dense video events.

\bibliographystyle{ACM-Reference-Format}
\balance
\bibliography{sample-base}

\end{document}